\documentclass[letterpaper, 10 pt, conference]{ieeeconf}  

\IEEEoverridecommandlockouts                              

\usepackage[utf8]{inputenc}
\usepackage{verbatim}
\usepackage{graphicx} 
\usepackage{epsfig} 
\usepackage{mathptmx} 
\usepackage{times} 
\usepackage{bm}
\usepackage{tikz}
\usetikzlibrary{shapes,arrows,positioning,fit}
\usepackage{mathtools}
 \usepackage{nccmath}
\usepackage[font=footnotesize]{caption}
\usepackage[font=footnotesize]{subcaption}
\usepackage{float}
\usepackage{amsmath} 
\usepackage{amssymb}  
\usepackage[ruled,vlined,linesnumbered]{algorithm2e}

\graphicspath{{./figs/}}
\usepackage{color}
\usepackage[normalem]{ulem}

\newtheorem{question}{RQ}
\let\labelindent\relax
\usepackage{enumitem}

\renewcommand{\baselinestretch}{0.965}

\newcommand\oprocendsymbol{\hbox{$\bullet$}}
\newcommand\oprocend{\relax\ifmmode\else\unskip\hfill\fi\oprocendsymbol}

\title{{Robot Navigation in Risky, Crowded Environments}: \\ Understanding Human Preferences}
\author{Aamodh Suresh*, Angelique Taylor**, Laurel D. Riek**, Sonia Mart\'inez*
 \thanks{This work was supported by grants XXX-XXXX-XXX}
 \thanks{* A. Suresh and S. Mart\'inez are with the Department of Mechanical and Aerospace Engineering, University of California San Diego, La Jolla, CA 92093, USA. E-mail: \{aasuresh,soniamd\}@eng.ucsd.edu. }
 \thanks{** A. Taylor and L. Riek are with the Department of Computer Science and
Engineering, University of California San Diego, La Jolla, CA 92093, USA. E-mail: \{amt062,lriek\}@eng.ucsd.edu. }
}

\begin{document}
\maketitle

\begin{abstract}
      Risky and crowded environments (RCE) contain abstract sources of risk and uncertainty, which are perceived differently by humans, leading to a variety of behaviors. 
     Thus, robots deployed in RCEs, need to exhibit diverse perception and planning capabilities in order to interpret other human agents' behavior and act accordingly in such environments. 
     To understand this problem domain, we conducted a study to explore human path choices in  RCEs, enabling better robotic navigational explainable AI (XAI) designs. 
     We created a novel COVID-19 pandemic grocery shopping scenario which had time-risk tradeoffs, and acquired users' path preferences. 
     We found that participants showcase a variety of path preferences: from risky and urgent to safe and relaxed. To model users' decision making, we evaluated three popular risk models (Cumulative Prospect Theory (CPT), Conditional Value at Risk (CVAR), and Expected Risk (ER). We found that CPT captured people's decision making more accurately than CVaR and ER, corroborating theoretical results that CPT is more expressive and inclusive than CVaR and ER. 
     We also found that people's self assessments of risk and time-urgency do not correlate with their path preferences in RCEs. 
     Finally, we conducted thematic analysis of open-ended questions, providing crucial design insights for robots is RCE. Thus, through this study, we provide novel and critical insights about human behavior and perception to help design better navigational explainable AI (XAI) in~RCEs. 
\end{abstract}

\section{Introduction}
\label{sec:introduction}

Robots are increasingly being deployed in everyday risky and crowded environments (RCE), including  shopping malls, museums, streets, and sidewalks (i.e.,~autonomous cars)~\cite{taylorsocial}.
These environments are often crowded, contain multiple sources of risk (e.g.,~dynamic and chaotic human-motion trajectories) and uncertainty (e.g.~noisy sensor measurements, including those from camera ego-motion \cite{taylor2022regroup}). As robots become more integrated into such environments, they need to appropriately deal with these challenges and navigate in a safe and socially-acceptable manner~\cite{taylorsocial,taylor2020situating,taylor2022hospitals,AS-SM:20}.

Modeling of how humans perceive risk~\cite{kwon2020humans} can help us understand and close this gap. These models differ on  the degree of rationality assumptions made on the human when subject to risky choices. These range from the consideration of human behavior  as completely rational and possibly risk-averse (e.g.,~Expected Risk (ER), Conditional Value at Risk (CVaR) ~\cite{RTR-SU:00}) to non-rational and possibly risk-insensitive (e.g.,~Cumulative Prospect Theory (CPT)~\cite{AT-DK:92}).  

However, little is known about the validity of these models in a risky social navigation setting, as well as how they compare with humans' self perception of risk. In particular, we are interested in understanding how robots can reason with humans, explaining their behaviors and actions, also known as Explainable Artificial Intelligence (XAI)~\cite{xu2019explainable}. 
XAI ``explains'' itself by opening up its reasoning to human scrutiny, resulting in better, faster, more accurate and more aligned human-robot decisions \cite{das2020opportunities,papagni2021understandable}.

\paragraph*{Prior Work} Risk is a relevant notion of urgency used to design navigation algorithms in robotics \cite{sharma2020risk}. 
Accordingly, various models have been employed to quantify and reason about risk. CVaR is one such popular model adopted from finance in robotics~\cite{choudhry2021cvar,sharma2020risk}, which captures risk aversion (i.e., ``play it safe'') by employing linear and rational notions of decision making. While this is analytically convenient, it 
cannot capture non-linear and non-rational decision making that humans usually exhibit~\cite{SSS:70,SSS:57,SD:16}.

\begin{figure}
    \centering
    \includegraphics[width=0.95\linewidth]{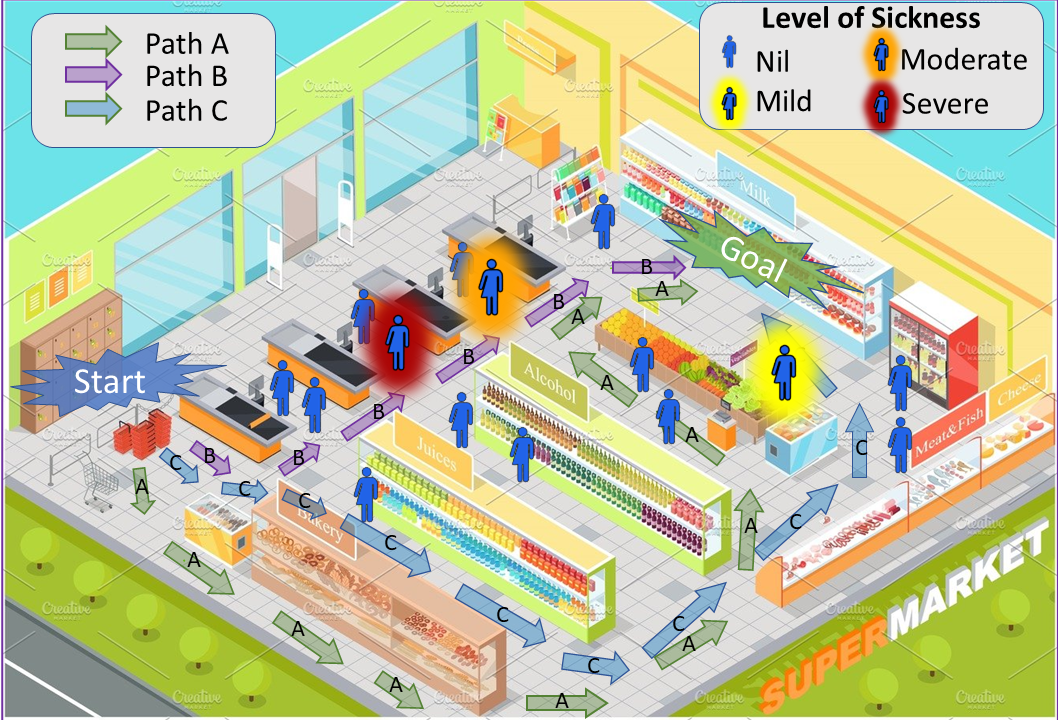}
    \caption[Grocery store environment]{Grocery store environment used in user studies. Participants select one of three paths (A, B, and C), to go from the entrance (shown as the `Start') to the milk section (shown as the `Goal'). The supermarket is crowded with people with levels of sickness ranging from `Nil' to `Severe'.}
    \label{fig:HRI_grocery_store}
    \vspace{-5mm}
\end{figure}

Recently, CPT methods~\cite{AT-DK:92} have been proposed~\cite{AS-SM:21-ral,AS-SM:21-lcss} to address this shortcoming. Theoretically, it has been shown that CPT is more ``expressive''~\cite{AS-SM:21-ral}, ``versatile'', and ``inclusive''~\cite{AS-SM:21-lcss} than CVaR and Expected Risk (ER), thus capturing a wider range of risk profiles of humans.  Preliminary evidence that CPT better captures human decision making under risk can be found in applications of traffic intersection management and routing~\cite{kwon2020cpt}, 
and resource management settings by operators~\cite{CPT-utility-user-study:20}. 
In practice, these approach is yet to be evaluated extensively in user studies pertaining navigation in RCE.

To do so, user studies that employ natural or explainable metrics to humans need to be developed. Unfortunately, commonly used risk variables such as money~\cite{AM-MP:17}, time~\cite{SG-EF-MBA:10}, or collision probabilities~\cite{hakobyan2020wasserstein}, do not satisfy this criterion for all cases. In fact, recent studies have found that humans are often sub-optimal in planning paths in such situations~\cite{reddy2018you}. These studies assume that the human is either ``noisy-rational'' or do not have correct environment models to choose optimally.

A few other avenues of using risk for planning paths in RCE include fall risk assessment~\cite{novin2020risk,koide2020collision,ballesteros2018automatic}, risk of localization and mapping systems \cite{hafez2019integrity,arana2020localization}, and planning risk in search and rescue operations \cite{shree2021exploiting}. 
These arguments are from a robot's perspective which acts in an expected manner and also expects the human to do so. 
However, from a human-centered and XAI perspectives, the robot's ``expected'' behavior might lead to mistrust and confusion~\cite{kruse2013human,shneiderman2022human}. To the best of our knowledge, general studies pertaining to everyday scenarios that employ more abstract cost interpretations are lacking, and are needed for better explainable AI design for robots in RCEs.


\begin{figure}[t] 
\centering 
\includegraphics[width=0.5\textwidth]{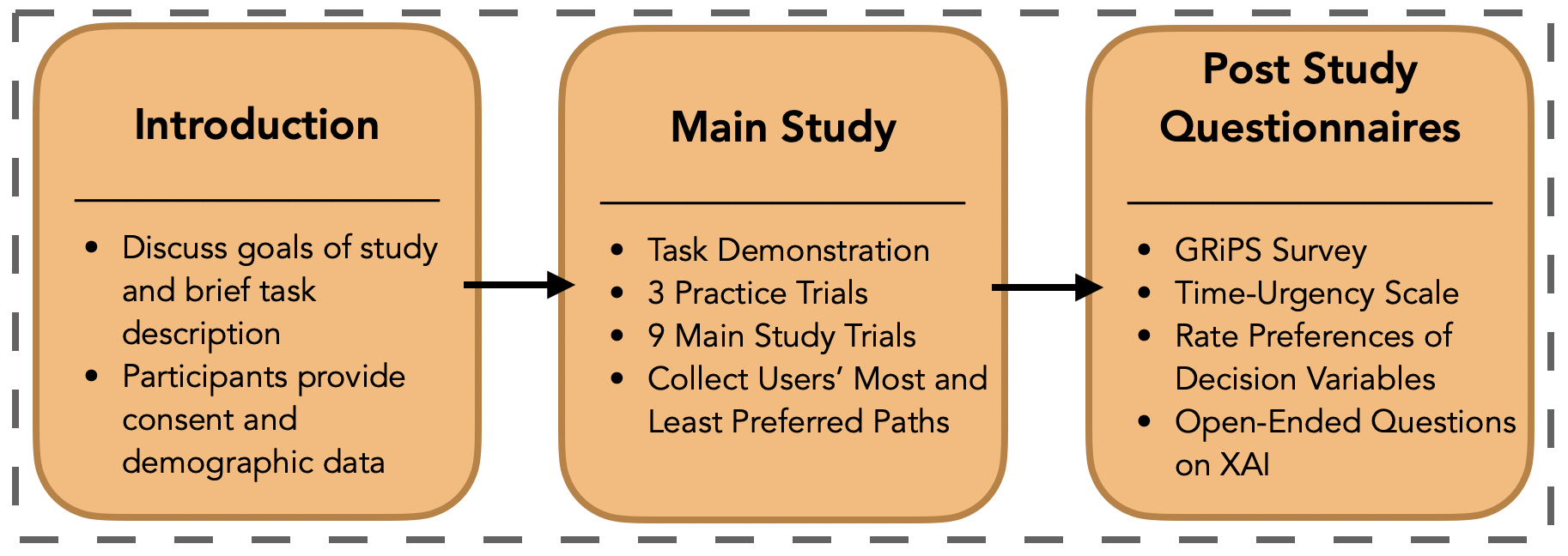} 
\caption{An overview of our study design.} 
\label{fig:overview}
\vspace{-5mm}
\end{figure}

\paragraph*{Contributions} In this work, through the design of a novel user study, we bridge the gap in existing literature by characterizing human perception of risk in RCEs, comparing theoretical risk models with observed human responses, and by exploring the consistency of human perception of risk and time urgency with standard survey responses. In addition, we provide new valuable insights for XAI design. 
Specifically, our work aims to address the following research questions:
\begin{question}
\label{prob:HRI_rq1}
What is the relationship between participants' path preferences and those arising from standard risk models?
\end{question}
\begin{question}
\label{prob:HRI_rq2}
What is the relationship between participants' self-risk and self-time-urgency perception and their actual path choices?
\end{question}
\begin{question}
\label{prob:HRI_rq3}
How do humans relatively weight time and risk to make navigational decisions in everyday scenarios?
\end{question}
\begin{question}
\label{prob:HRI_rq4}
What are the users' preferences to interact with robots navigating in everyday scenarios?
\end{question}
We conducted a large scale online study (n = 82) and found that most participants do not make decisions in an expected manner (in accordance with expected risk metric) and that CPT as a risk model captures the observed responses better than CVaR and ER.
Interestingly, through the application of standard questionnaires, we find that there is a mismatch between humans' self-risk/self-time-urgency assessment and their actual choices. Additionally, participants generally give a higher weight to risk than time while choosing paths. 

Finally, we provide valuable insights to design XAI for robots in  RCEs. For example,  we found that most participants want robots that can explain its rationale behind decision-making and they also suggested user interface design to have a two-way motion intention communication between users and robots. Thus, equipped with these results and insights, XAI design can be improved to enable robots to operate and adapt to human preferences in RCEs. 
\section{Methodology}
\label{sec:HRI_methodology}
  



 We conducted an IRB-approved (approval code: 201638) within-subjects study on the Qualtrics\footnote{https://www.qualtrics.com/} survey platform. The study was designed  to the evaluate people's risk perception, and was inspired by the COVID-19 pandemic. This provides an easy-to-relate context for participants to think about decision making under risk. 
Thus, we consider a grocery-store shopping scenario, where the risk is characterized by being coughed at by potentially infected people. 
Participants were asked to imagine being an ``Instacart\footnote{A grocery delivery service (\texttt{www.instacart.com}).} Shopper'' who needs to go from the entrance of the store to the milk section. Time-urgency is characterized by the need to complete shopping quickly in order to get better ratings and tips.

This scenario is illustrated in Figure~\ref{fig:HRI_grocery_store}. Here, participants had three paths to choose from.  Each path had a varying intensity of risk and time urgency (discussed in detail in Section~\ref{sec:hri_scenario_logic}). 
The participants indicated their most and least preferred paths for each scenario.
In the following paragraphs, we explain the scenario methodology and the list the post-study questions that we use (see Tables~\ref{tab:HRI_GRiPS} and~\ref{tab:HRI_time}). 

\paragraph*{Participants}
\label{sec:evaluation}
We recruited 82 participants affiliated with a university campus through university list-serves and via word of mouth. The participants consisted of 27 females, 49 males, 1 binary/third gender and 5 that preferred not to answer this question. The ages ranged from 21-32 (mean = 25.6, SD = 2.5) and their educational background had a distribution of 68 in Engineering, 3 in Mathematics, 5 in Basic Sciences, 1 in Management, and 5 from other fields.

\subsection{Measures and metrics employed}
\label{sec:post_study_questions}
We obtained the following measures and metrics from each participant through nine trials and post-study questionnaires:


\subsubsection{Path preferences}
For each trial the participant revealed their path preference order by providing their most preferred path (MPP) and Least preferred path (LPP). The MPP is denoted as $ M \triangleq \{m_1 ,m_2, ... , m_9\}$, for all 9 trials. For a trial $i$, we encode the user's MPP choice as $m_i\triangleq 0$ for path A (resp.~$m_i\triangleq 1$ if they chose path B, and $m_i\triangleq 0.5$ if they chose C), indicating the level of relative risk and time-urgency. 
Similarly the Least preferred path (LPP) is denoted as $ L \triangleq \{l_1 ,l_2, ... , l_9\}$, for all 9 trials. We encode $l_j\triangleq 0$ for path A (resp.~$l_j\triangleq 1$ if they chose path B, and $l_j\triangleq 0.5$ if they chose C), indicating the level of relative risk and time-urgency). Using these measures we describe the risk taking behavior of the participants.
\subsubsection{Observed risk taking behavior}
We characterize participants' decision making into three different categories based on the risk taking behaviors expressed. First, we consider ``expected behavior'' using Expected Risk (ER), from which we get $M^{exp}$ and $L^{exp}$. These are the MPP and LPP choices when risk is defined in an expected manner. Next, we observe ``risk aversion'' through CVaR, where we similarly obtain $M^{av}$ and $L^{av}$. Finally, we consider ``risk insensitive" behavior using CPT with $M^{ins}$ and $L^{ins}$ as the MPP and LPP choices,  respectively. We note that from our previous theoretical results~\cite{AS-SM:21-ral,AS-SM:21-lcss}, we have shown that CPT is the most inclusive model and can capture all three perceptions. CVaR can capture expected and risk averse perception, whereas ER only captures the expected behavior. In the following, we list the corresponding metrics, and how they are calculated.

\begin{enumerate}[label=\alph*)]
    \item Average MPP score, $\overline{M}$: This is the average risk and time-urgency of the participants' MPP, $ \{m_1 ,m_2, \dots , m_9\}$.
    \item Average LPP score, $\overline{L}$: This is the average risk and time-urgency of the participants' LPP, $ \{l_1 ,l_2, \dots , l_9\}$.
    \item Deviation from expected behavior, $J^{exp} = \sum_{i=1}^9 |m_i-m^{exp}_i|$: Larger values indicate greater deviations from the expected behavior. 
    \item Deviation from risk aversion, $J^{av} = \sum_{i=1}^9 |m_i-m^{av}_i|$: Larger values indicate greater deviations from risk-averse (CVaR) behavior.
    \item Deviation from risk insensitivity, $J^{ins} = \sum_{i=1}^9 |m_i-m^{ins}_i|$: Larger values indicate greater deviations from risk insensitive behavior.
    \item Deviations from RPMs: For ER we have $J^{ER}=J^{exp}$. For CVaR we have $J^{CVaR}=\min \{J^{exp},J^{av}\}$ and for CPT we have $J^{CPT}=\min \{J^{exp}, J^{av}, J^{ins}\}$.
\end{enumerate}
Next, we look at the self-reported measures that the users provide us through post-study questionnaires.

\subsubsection{Self-reported measures}
After collecting user path choices, we then conducted these post-study questionnaires.

\textbf{GRiPS:} General Risk Propensity Scale (GRiPS)~\cite{zhang2019development}, which measures the participants' self risk-taking abilities, i.e. it evaluates how risk-averse or risk-taking they think they are in their daily lives.
GRiPS is a self-report measure (see Table \ref{tab:HRI_GRiPS}) of general risk and pro-social behavior consisting of 8-items which participants answer on a Likert scale from 1 (Strongly Disagree) to 5 (Strongly Agree). We denote the responses for the GRiPS questionnaire as $R \triangleq \{r_1, r_2 , ... , r_8\}$ for the $8$ questions.
\begin{table}[t]
\centering
\caption{The 8-item the General Risk Propensity Scale (GRiPS) ~\cite{zhang2019development} that we administered to participants after engaging in our study.}
\label{tab:HRI_GRiPS}
\begin{tabular}{l}
\hline
\textbf{GRiPS Survey Questions}                              \\ \hline
1. Taking risks makes life more fun                      \\
2. My friends would say that I'm a risk taker            \\
3. I enjoy taking risks in most aspects of my life       \\
4. Taking risks is an important part of life             \\
5. I commonly make risky decisions                       \\
6. I am a believer of taking chances                     \\
7. I would take a risk even is it meant I might get hurt \\
8. I am attracted, rather than scared, by risk           \\ \hline
\end{tabular}
\vspace{-5pt}
\end{table}

\textbf{Time Urgency:} The second questionnaire is called the ``Time Urgency Scale''~\cite{FJL-HR-JT-CC:91}, which measures participants' self assessment of how time-urgent and urgent they think they behave in everyday scenarios.
It is a self-report measure (See Table~\ref{tab:HRI_time}) of general time-related behavior consisting of 6-items (as commonly used~\cite{mohammed2011temporal}), which participants answer on a Likert scale from 1 (Strongly Disagree) to 5 (Strongly Agree). We denote the responses for the Time-Urgency questionnaire as $T \triangleq \{t_1, t_2 , ... , t_6\}$ for the $6$ questions.
\begin{table}[t]
\centering
\caption{The 6-item the Time Urgency Scale~\cite{FJL-HR-JT-CC:91} that we administered to participants after engaging in our study.}
\label{tab:HRI_time}
\begin{tabular}{l}
\hline
\textbf{Time Urgency Survey Questions}                              \\ \hline
1. I find myself hurrying to get places even when there is plenty of time. \\
2. I often work slowly and leisurely. \\
3. People that know me well agree that I tend to do most things in a hurry.\\
4. I tend to be quick and energetic at work.\\
5. I often feel very pressed for time.\\
6. My spouse or a close friend would rate me as definitely relaxed \\ and easy going.\\ \hline
\end{tabular}
\vspace{-15pt}
\end{table}

We employ metrics to evaluate the participants' self perception of risk and their observed risk-taking behavior, which are described next.

\begin{enumerate}[label=\alph*)]
    \item \textit{Risk score $\overline{R}$}: The average response (normalized between $0$ and $1$) from the GRiPS survey responses $R$ (as commonly used~\cite{GRiPS_use:20}). Here, a $\overline{R}=1$ indicates an adventurous perception, whereas $\overline{R}=0$ indicates a risk-averse perception. 
    \item \textit{Time urgency score $\overline{T}$}: The average response (normalized between $0$ and $1$) from the Time-urgency survey responses $T$. Here a $\overline{T}=1$ indicates a hasty behavior, whereas $\overline{T}=0$ indicates a relaxed behavior. 
    \item Risk similarity score $R^{sim}=\overline{M}- \overline{R}$: measures the deviation from users' self-risk perception and their observed risk perception. $R^{sim} \in [-1,1]$, where $R^{sim} \approx 0$ indicates user's self-perception of risk and their observed risk taking characteristics are similar. A larger positive (negative) value indicates users chose riskier (safer) paths than their self perception through the questionnaire.
    \item Time-urgency similarity score ,$T^{sim}=\overline{M}- \overline{T}$: Shows if people choose paths according to the time-urgency survey responses. As above,  a $T^{sim} \in [-1,1]$, $T^{sim} \approx 0$, indicates similarity in user's self perception and observed behavior of time urgency.  Larger positive (negative) values indicate users choose more urgent (leisured) paths than their self perception.
    \end{enumerate}
\textbf{Decision Variable Preferences:} In addition, after the study trials and questionnaires are administered, we ask participants how they relatively weighed (as a \%) each of the four variables in making their decisions:  that is, the time taken, the number of sick people, the level of sickness, and the chance of being coughed at for a particular path. The participants' relative reliance (in \%) on each decision variable is denoted by $\{v_1, v_2 , ... , v_4\}$ corresponding (Table~\ref{tab:HRI_decision_variable}) to the four variables. 

\subsubsection{Open ended questions}

We finally asked open-ended questions (listed below) to better understand human preferences towards designing robot navigation in RCE. 
\begin{itemize}
    \item Q1: Would you like to know how robots make decisions and plan paths? If so, how do you want a robot to explain its thought processes behind its decisions and what modality (e.g., speech, expressions) would you prefer? 
    \item Q2: How do you want the robot to communicate its movement intentions (e.g., moving right or left)? 
    \item Q3: Would you like a robot to know how you are making decisions and planning paths? If so, how do you want to explain your intentions and what modality (e.g., speech, touchscreen) would you choose?
\end{itemize}

To summarize, the trial data helped us answer RQ~\ref{prob:HRI_rq1}. 
Question RQ~\ref{prob:HRI_rq2} can be assessed from analyzing the trial data along with the questionnaire responses. The data on decision-variable preferences helped us answer RQ~\ref{prob:HRI_rq3}. Whereas the responses to open-ended questions helped us answer RQ~\ref{prob:HRI_rq4}.



\subsection{Scenarios}
\label{sec:hri_scenario_logic}
Participants were presented with three choices of paths to choose from, including paths A, B, and C (see Figure~\ref{fig:HRI_grocery_store}).
Path A was the longest, path B was the shortest, and path C had a length that was in between.
We used the situation of ``being coughed at by sick people'' to elicit risk for each path in every scenario. This risk was described by four decision variables: ``time taken'', ``number of sick people'', ``level of sickness'', and ``chance of being coughed at'' (see Table~\ref{tab:HRI_decision_variable}). The time taken varied from 5 to 20 minutes, the number of sick people from $0$ to $2$, the level of sickness from $0$ to $3$, and the chance of being coughed at was expressed as a percentage for each sick person encountered. Regarding the level of sickness, we used the following terminology: 0-Nil, 1-Mild, 2-Moderate and 3-Severe (see Figure~\ref{fig:HRI_grocery_store}). We purposefully kept the consequences of being exposed to a sick person abstract, in order to extract realistic risk perceptions from participants. The variables are summarized in Table~\ref{tab:HRI_decision_variable}. 

We administered nine trials with different values for decision variables, aimed at capturing a wide range of scenarios.  In each trial, the shortest path (Path B) had the most risk and uncertainty, while the longest path (Path A) had the least risk and uncertainty. Participants choose their most and least preferred paths, thus providing a preference order. The risk variables for each trial are designed in such a way that the best expected choice\footnote{That is, best according to the expected risk metric.}  (A or B or C) varies across trials. We then group the nine trials into three levels of uncertainty (w.r.t. number of sick people). Comparing the chosen preferences with those based on expected risk, and across different levels of uncertainty, helps us understand how human choices are affected by uncertainty. 

\begin{table}[]
\centering
\caption{Description of decision variables and their ranges for each path in every scenario.}
\label{tab:HRI_decision_variable}
\begin{tabular}{|l|l|l|}
\hline
\textbf{ No.} & \textbf{Decision Variables} & \textbf{Range of values presented}                                                            \\ \hline
1                & Time Taken                  & \begin{tabular}[c]{@{}l@{}}Path A : 20 mins\\ Path B : 5 mins\\ Path C : 10 mins\end{tabular} \\ \hline
2                & Number of Sick people       & \begin{tabular}[c]{@{}l@{}}Path A : 0-1\\ Path B : 2-3\\ Path C : 1-2\end{tabular}            \\ \hline
3                & Level of Sickness           & 0-3 for each path                                                                             \\ \hline
4                & Chance of being coughed at  & 0-100 \% for each path                                                                        \\ \hline
\end{tabular}
\vspace{-5mm}
\end{table}


\subsection{Procedure}
An overview of our study design is available in Figure~\ref{fig:overview}. The study began with a consent form providing a brief description of the study. After giving informed consent, participants provided their demographic information including age, gender, occupation, and area of expertise. The main study then started, with a discussion the goal of the study, which is to investigate how people choose paths in risky situations. We further explained the user interface (UI) and elements of the study through a demonstration trial.

Participants then saw a demonstration scenario of the grocery store (Fig.~\ref{fig:HRI_grocery_store}), describing the four decision variables pictorially, in sentences and through a summary table\footnote{These decision variables were selected after a few rounds of pilot studies which indicated different people prefer different variables to describe the scenarios.} (see Table~\ref{tab:HRI_decision_variable}). Then, they selected their most and least favorite paths. Based on their most preferred path choices, we randomly selected a risk outcome and display the final results\footnote{Through pilot studies we learned that displaying the trial results enhanced user engagement.} (e.g., `You encountered no sick people'). 

Next, participants engaged in three practice rounds with three different scenarios and selected their most and least favorite paths. After the practice rounds, they then participated in the main study, which presented 9 different combinations of  ``risk'' in each scenario. To remove ordering effects and the influence of regret, we randomized the ordering of the trials across all participants.

At the conclusion of the study, we administered the GRiPS ~\cite{zhang2019development} and Time-Urgency~\cite{FJL-HR-JT-CC:91} questionnaires, to respectively measure self risk-taking and time-urgency perception among participants. In addition, we asked participants to relatively weigh (as a percentage of) each of the four variables to make their path choices. 
Finally, we asked open-ended questions (listed in Section ~\ref{sec:post_study_questions}) to better understand their preferences towards designing explainable AI for path planning in everyday scenarios.

\subsection{Analysis} 
\label{sec:qual_analysis}

\textbf{RQ1 Analysis: } We examined the relationship between participants' path choices and the the corresponding choices that result from a risk model. To do this, we analyzed the descriptive statistics of the objective metrics that measure deviation from various behaviors and risk models. (As we had only 9 trials, the sample was insufficient for parametric correlation analysis.)

\textbf{RQ2 Analysis:} We explored the relationship between participants self-perception of risk and time-urgency, compared to the expected (baseline) risk associated with their path choices in terms of most and least preferred paths. In this regard, we  analyzed the descriptive statistics of the subjective metrics ($\overline{R}$ and $\overline{T}$) and the similarity scores ($R^{sim}$ and $T^{sim}$). We  also conducted a correlation analysis to understand the interaction between self-perception of risk and time-urgency. We also performed a pairwise comparison of means (paired t-test) to reveal additional trends between variables representing participants choices and self perception.

\begin{figure}
    \centering
    \includegraphics[width=0.99\linewidth]{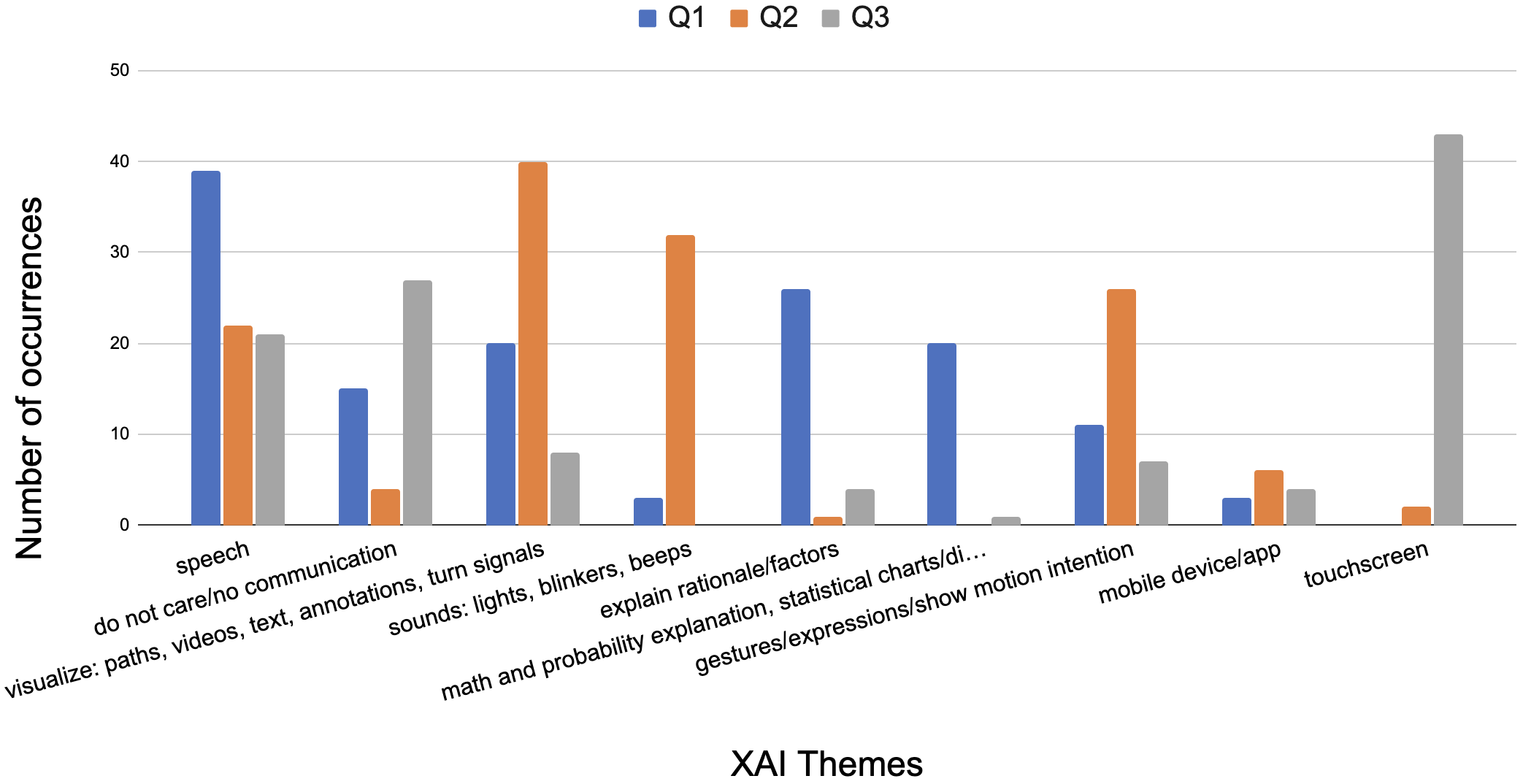}
    \caption[Open-ended question statistics]{This shows the number of occurrences of each explainable AI (XAI) theme identified in the data collected in responses to three open-ended questions from Section \ref{sec:post_study_questions} to address RQ4. }
    \label{fig:open_ended_question_stats}
    \vspace{-5mm}
\end{figure}

\textbf{RQ3 Analysis}: We studied the relative importance that users give to each decision variable (in Table~\ref{tab:HRI_decision_variable}) in order to choose their most preferred and least preferred paths. To do this, from this data, we  created two new variables to measure the relative importance of time and risk used to make decisions. This was done by first averaging the three variables representing risk, and then expressing it as a percentage w.r.t.~the total of time taken and average risk percentages. We provide descriptive statistics on the responses to post-study question on decision variable preferences.

\textbf{RQ4 Analysis: } We determined user preferences for XAI design through open-ended questions. Two members of our team performed thematic coding on open-ended question responses following grounded theory~\cite{braun2012thematic} as commonly done in the literature. This process involved reviewing responses (see Section~\ref{sec:post_study_questions}), generating high-level themes to capture key ideas in the data, reviewing the thematic codes with the team, negotiating them based on key ideas in the data, and repeating the process until all codes have been agreed upon. Next, we coded a total of 246 codes with our final set of codes shown in the x-axis of Figure \ref{fig:open_ended_question_stats}. We computed the inter-rater agreement using Krippendorffs-Alpha as we used multiple codes for each quote. This is advantageous because it supports categorical and ordinal data. We found an IRR of 1.0 which is considered high agreement. We believe this is due to sparse responses with 2126 words total across 82 participants. There was an average of 8.64 words per response, a minimum of 1, a maximum of 102, and a median of 4 words per response across the dataset.

\section{Results}
We provide descriptive statistics of the relevant variables and metrics, including the mean, median, standard deviation, and 95\% confidence interval.
To study the correlation between two variables, we calculate the  Pearson's correlation coefficient, along with null-hypothesis significance testing with threshold p-value = $0.05$.


\subsection{RQ1: Comparing users' risk perception}

In order to compare risk models with users' decision-making, we provide descriptive statistics of the relevant variables, which are summarized in Table~\ref{tab:HRI_rq1_descriptive}. Recall that MPP (similarly, LPP) for a $j^{th}$ trial with $m_j=0$ is path A with lowest risk and most leisured. Whereas $m_j=1$ is path B with the highest risk and most time-urgency, and Path C is in between in both risk and time-urgency with a value of $m_j=0.5$.

\begin{table}[]
\centering
\caption{Descriptive statistics of variables to compare users' decisions with standard risk-model decisions to address RQ~\ref{prob:HRI_rq1}.}
\label{tab:HRI_rq1_descriptive}
\resizebox{\linewidth}{!}{%
\begin{tabular}{|c|c|c|c|c|}
\hline
\textbf{Variable} & \textbf{Mean} & \textbf{\begin{tabular}[c]{@{}c@{}}95\% confidence\\ interval of mean\end{tabular}} & \textbf{\begin{tabular}[c]{@{}c@{}}Standard\\ Deviation\end{tabular}} & \textbf{Range} \\ \hline
$\overline{M}$    & 0.60                       & 0.56 to 0.66                                                                        & 0.20                                                                   & 0.00 to 1.00            \\ \hline
$\overline{L}$    & 0.40                      & 0.35 to 0.47                                                                        & 0.30                                                                   & 0.00 to 0.94         \\ \hline
$J^{exp}$ for MPP     & 0.40                       & 0.35 to 0.40                                                                        & 0.10                                                                   & 0.11 to 0.61         \\ \hline
$J^{av}$ for MPP     & 0.30                      & 0.31 to 0.36                                                                        & 0.10                                                                   & 0.00 to 0.61         \\ \hline
$J^{ins}$ for MPP     & 0.30                       & 0.28 to 0.36                                                                        & 0.20                                                                   & 0.00 to 0.89         \\ \hline
$ J^{CVaR}$ for MPP    & 0.30                       & 0.31 to 0.36                                                                        & 0.10                                                                   & 0.00   to 0.61         \\ \hline
$J^{CPT}$ for MPP     & 0.20                       & 0.21 to 0.26                                                                        & 0.10                                                                   & 0.00 to 0.44         \\ \hline
\end{tabular}%
}
\vspace{-5mm}
\end{table}

\textbf{Path choice characteristics:}
From Table~\ref{tab:HRI_rq1_descriptive}, the mean, median, and confidence interval are over 0.5 for the average $\overline{M}$, and under 0.5 for the average $\overline{L}$. This indicates a preference towards Path B (more risky and time urgent), and a disinclination towards Path A (more safe and leisured). Additionally, $\overline{M}$ exhibits a full range from $0-1$, whereas $\overline{L}$ has a range $0-0.94$. Thus, although their preferences were scattered across the full spectra, no participant disliked path B across all trials.

\textbf{Behavioral Characteristics and model comparisons:}
From Table~\ref{tab:HRI_rq1_descriptive}, we note that the deviation from the expected behavior $J^{exp}$ is in general greater than $J^{av}$ and $J^{ins}$, indicating that expected behavior (from expected risk) is the least aligned with participants' preferences. Also, from row 3, the min is $>0$, indicating that not a single participant showed expected behavior across all their trials.  
The almost-similar statistics for risk-averse and risk-insensitive behaviors show that participants' exhibited both of these behaviors equally frequently. Also, the deviation $J^{CPT}$ is the least, showing that CPT is a better model to approximate the participants' decision making.
Similar deviation statistics were correspondingly obtained for LPP choices; hence, we omit its discussion here. 

\subsection{RQ2: Users' Self-Perception of Risk vs. Expected Risk}
We provide descriptive statistics of the relevant variables (Table~\ref{tab:HRI_rq2_descriptive}) and conduct correlation studies next. From Table~\ref{tab:HRI_rq2_descriptive}, the statistics of the average survey responses $\overline{R}$ indicate that, in general, participants are more inclined towards taking risks, as the mean, median, and confidence interval are all over 3. 
However, a wide range and high standard deviation suggest a fairly diverse set of risk-taking behaviors. A similar trend is observed for time-urgency $\overline{T}$.
To compare the participants' decision making characteristics with their self perception of time-urgency and risk, we first consider the similarity scores $R^{sim}$ and $T^{sim}$ (Table~\ref{tab:HRI_rq2_descriptive}). The risk similarity score $R^{sim}$ for MPP and LPP is balanced with the mean, median, and confidence interval close to 0, but have a high range and standard deviation. Hence, there are  a variety of people with different perceptions, and the GRiPS responses may not fully represent their decision making in the study. A similar trend is observed for time urgency, with a slightly greater inclination of participants to act more urgently than what they indicate in the survey.

\begin{table}[]
\centering
\caption{Descriptive statistics of similarity scores of users' decisions in trials compared to their questionnaire responses to address RQ~\ref{prob:HRI_rq2}.}
\label{tab:HRI_rq2_descriptive} 
\resizebox{\linewidth}{!}{%
\begin{tabular}{|c|c|c|c|c|}
\hline
\textbf{Variable} & \textbf{Mean} & \textbf{\begin{tabular}[c]{@{}c@{}}95\% confidence\\ interval of mean\end{tabular}} & \textbf{\begin{tabular}[c]{@{}c@{}}Standard\\ Deviation\end{tabular}} & \textbf{Range} \\ \hline
$\overline{R}$    & 0.6                       & 0.55 to 0.64                                                                        & 0.2                                                                   & 0.00 to 1.00            \\ \hline
$\overline{T}$    & 0.5                       & 0.48 to 0.56                                                                        & 0.2                                                                   & 0.08 to 0.96         \\ \hline
$R^{sim}$ for MPP      & 0.10                & -0.05 to 0.09                                                                       & 0.30                                                                   & -0.84 to 0.94         \\ \hline
$R^{sim}$ for LPP      & 0.00                         & -0.08 to 0.07                                                                       & 0.30                                                                   & -0.91 to 0.94         \\ \hline
$T^{sim}$ for MPP      & 0.10                    & 0.04 to 0.15                                                                        & 0.30                                                                   & -0.58 to 0.67         \\ \hline
$ T^{sim}$ for LPP     & 0.10                    & 0.00 to 0.14                                                                         & 0.30                                                                  & -0.58 to 0.67         \\ \hline
\end{tabular}%
}
\vspace{-10pt}
\end{table}

Next, we try to identify trends between the GRiPS and Time-Urgency survey responses and path choices by measuring correlation and performing linear regression. We highlight the significant results here. There was a significant interaction between the average survey responses for risk similarity score $R^{sim}$ and time urgency similarity score $T^{sim}$ for MPP and LPP with  and p $<$ 0.05 and a effect size \textless $0.5$.
This implies that people who acted more/less riskier than they indicated in the GRiPS survey, also acted correspondingly more/less time urgent than they indicated in the time-urgency survey. This can arise because of the study construction, where paths which are shorter (time urgent) are also riskiest and vice versa. 

Interestingly, there was an insignificant interaction between the normalized GRiPS response $\overline{R}$ with MPP and LPP scores $\overline{M}$/$\overline{L}$ with p = $0.936$ and p = $0.655$, and effect size close to $0$.
This reveals that participants' self risk evaluation and their path choices are not related. The trend was similar in the case of time-urgency. Thus, relying solely on the GRiPS and/or time-urgency survey to depict participants' behavior may not be effective. 

Table~\ref{tab:HRI_rq2_correlation_pd} shows the paired t-test statistics of various pairwise means comparisons. Interestingly we found significant evidence that $\overline{L}$ is greater than $\overline{R}$ (p <0.001) with a medium effect size (d > 0.5). On average, $\overline{L}$ is greater than $\overline{R}$ by 0.18 (95\% CI: 0.1, 0.26) units.
By this, we can infer that participants had more disinclination towards riskier  paths than their indicated risk appetite through the GRiPS survey. 

\subsection{RQ3: Users reliance on decision variables}

\begin{figure*}
     \centering
     \begin{subfigure}[b]{0.3\textwidth}
         \centering
         \includegraphics[width=\textwidth]{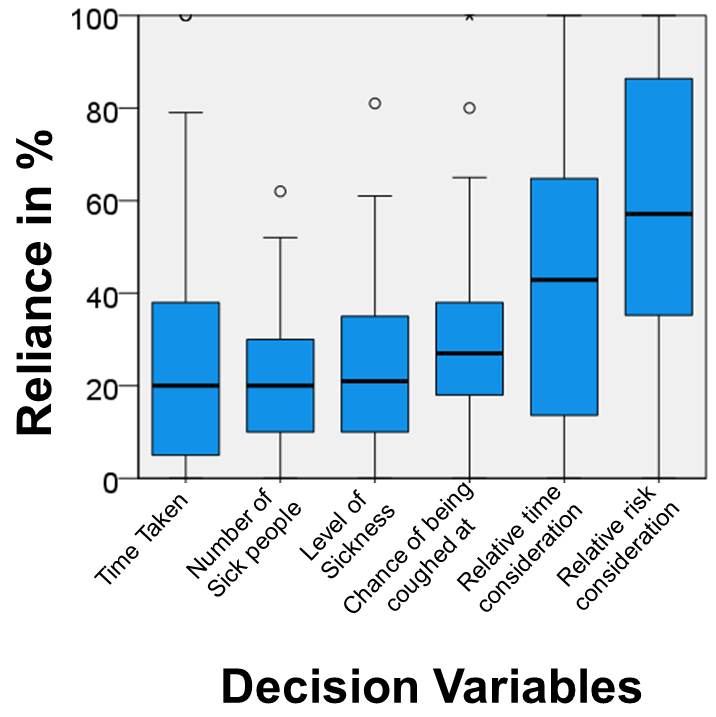}
         \caption{Reliance on decision variables}
         \label{fig:percentage_reliance}
     \end{subfigure}
     \hfill
     \begin{subfigure}[b]{0.68\textwidth}
         \centering
         \includegraphics[width=\textwidth]{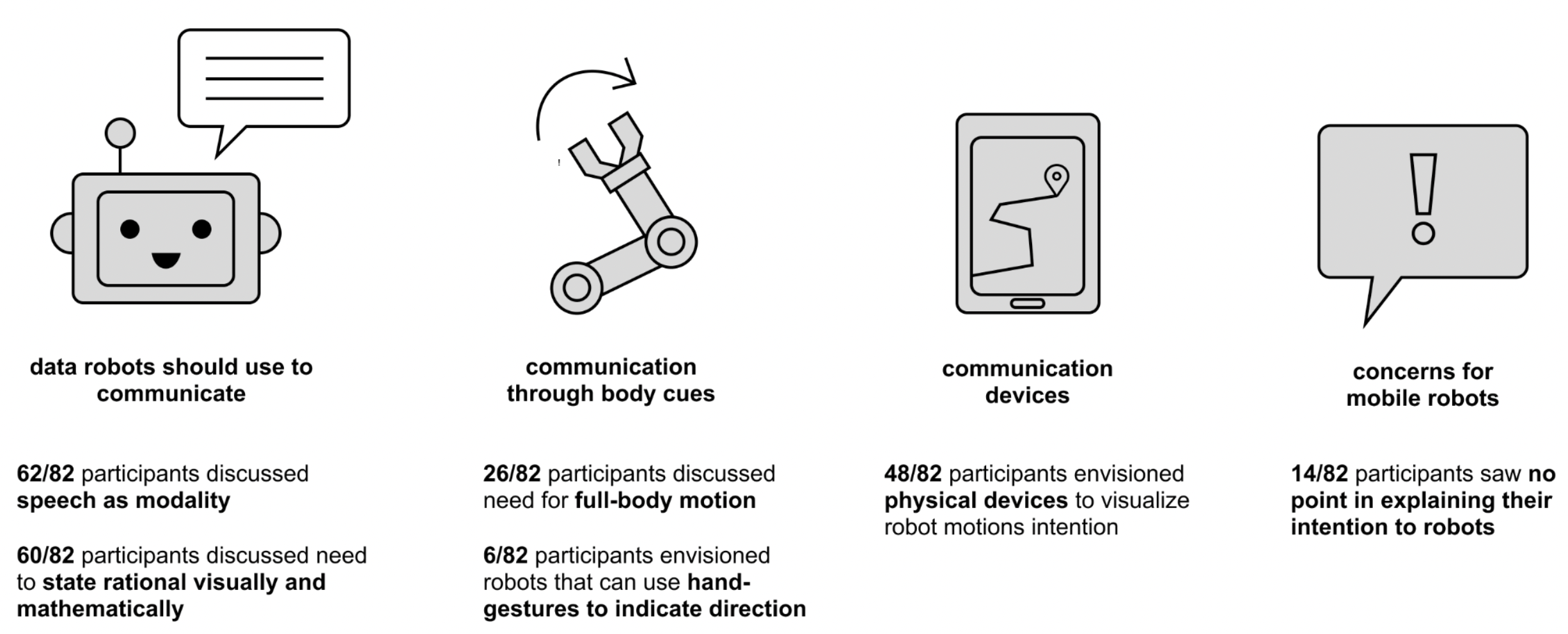}
         \caption{XAI Themes}
         \label{fig:open_ended_question_stats2}
     \end{subfigure}
        \caption{(a) Boxplots showing the users' preferences of decision variables (Table~\ref{tab:HRI_decision_variable}) in percentage to choose paths in the trials to address RQ~\ref{prob:HRI_rq3}. (b) Themes that inform users’ preferences for XAI systems in social navigation settings from Section \ref{sec:post_study_questions} to address RQ~\ref{prob:HRI_rq4}}
        \label{fig:open_ended_questions}
        \vspace{-5mm}
\end{figure*}

We measured the relative importance that participants give to the four decision variables (Table~\ref{tab:HRI_decision_variable}) to choose paths in terms of percentages. The data is described as boxplots (first 4 from the left) in Figure~\ref{fig:percentage_reliance}. The last two boxplots represent the relative time and risk consideration, respectively. We note that the relative consideration of time has a mean $40.8\%$ and median $42.8\%$, as opposed to relative consideration of risk which has mean $59.2\%$ and median $57.2$. So, in general, participants seem to consider risk factors more importantly than time factor while making decisions. However, with a large standard deviation of $30.2$ and a full range of $0-100$ for both variables, the generalization may not apply to many participants. This again reflects the diversity  regarding time and risk consideration for making path planning decision by humans. Thus, XAI needs flexible models in this regard.

\subsection{RQ 4: Users' Interaction Preferences}

We asked participants three open-ended questions from Section~\ref{sec:post_study_questions} to gauge their preferences w.r.t.~how they would like robots to communicate their intentions. 
After conducting the analysis as described in Section~\ref{sec:qual_analysis}, we found many redundant responses for the three questions. 
Thus, we discuss descriptive statistics for each question and describe the overarching themes we found in all responses.

\textbf{Descriptive statistics of open-ended questions: } We are interested in learning about users’ preferences for XAI-capable robots in everyday settings.
However, we found it challenging to fully address our research questions without also collecting qualitative data about users' preferences. 
Thus, we asked the open-ended questions from Section \ref{sec:qual_analysis}.

Figure~\ref{sec:post_study_questions} shows the descriptive statistics of the XAI themes used to code the data across all open-ended questions.
In summary, there are ten XAI themes identified in the data (x-axis of Figure~\ref{sec:qual_analysis}).
From here, 73/82 participants indicated interest in XAI systems, and 8/82 participants did not want robots that explain their motion intentions. On the other hand, 
69 out of 82 participants wanted robots that understand how users in their environment plan paths, 10 out of 82 do not, and 3 out of 82 participants indicated maybe. The ‘touchscreen’ theme achieved the highest occurrence count.
Several themes achieved the lowest occurrence score of 0 which include themes `sounds, lights, blinkers, beeps' for how the robot knows what users want, `math, probability, explanation, statistical charts/diagrams' for how to communicate intent, or `touchscreen' in terms of how they would like robots to plan paths.

\textbf{Thematic coding of Open-Ended Questions:} We identified four overarching themes that inform users' preferences for XAI robotic navigational systems.
These themes emphasize the importance of data variables used to interact with XAI systems for mobile platforms, how robots should communicate their navigation plans, and users' concerns about using XAI for mobile robots in everyday environments.

\textit{What data robots should use to communicate with users:} Users' identified a range of preferences for data robots can use to communicate its motion intentions.
The most popular modality discussed by 62 out of 82 participants was speech, as it is convenient to communicate naturally.
More specifically, 60 out of 82 participants discussed the need to state rationale visually and mathematically such as with pie charts, using arrows, or visual representation of weights used in AI decision-making.
Additionally, 26 out of 82 participants envisioned robots that can explain the factors behind their decision, consistent with prior work in XAI \cite{das2020opportunities,papagni2021understandable} e.g.,
level of sickness during COVID-19.
Lastly, participants discussed the need for mobile robots that adapt to users' movement over time.
P39 stated, ``I would want to know what the robot is `thinking' so that I know how to adjust my own behavior/path/location.'' 

\textit{Robots that communicate motion intentions with body cues: }Our analysis of the open-ended questions indicate that participants envision robots that communicate non-verbally using bodily cues.
For instance, 26 out of 82 users discussed the need for full-body motion.
They discussed robots that preemptively gesture in the direction they plan to move in before doing so.
Also, 6 out of 82 participants envisioned robots that can provide hand gestures to indicate the direction they plan to move in.
P84 said, ``Something like a turn indicator on a car. It should flash/get attention visually and make a noise if possible so that people with low vision/hearing would be able to notice it''.

\textit{Preferred communication devices with robots:} Our results show that users preferred several devices for communicating with robots.
Participants envisioned intelligent user interfaces to facilitate interaction with robots inspired by Google Maps as a top-down map of the robot and people around it in real-time. 
Building on this, they discussed a feature that enables them to view a ranked list of paths the robot is considering and why it chose its current path in terms of factors.
P9 said, ``I would like to know the top options the robot considered and why it decided to go with its final choice.''
Another idea was for robots to have `car-like' features like turn signaling with lights or using a human-actuated hand to point or gesture in the direction the robot intends to move in. 
Lastly, 48 out of 82 participants envisioned physical devices to visualize robot motion intention information including on their phone or touchscreen devices.

\textit{Concerns about XAI for mobile robots:} 14 out of 82 participants saw no point in explaining their intentions to robots.
Instead, they envisioned robots that adapt to them.
Furthermore, they wanted robots to be passive actors in their environment instead of an active agent that they can interact with, indicating ``[...] this should be inferred. I don't want to change my natural behavior'' -P21.
One salient reflection resulting from the analysis is that some users expressed concern for robots using speech to communicate and they foresaw it as `creeper' or `annoying'. 
P9 said, ``Speech is a convenient way [...], but it could get annoying if the robot is constantly talking about why it's going where it's going.''
Overall, there were a range of preferences identified in the data which highlights the need for systems with diverse XAI capabilities that adapt to users' preferences over time.

\begin{table}[t]
\centering
\caption{Paired t-test statistics of relevant variables to address RQ2}
\label{tab:HRI_rq2_correlation_pd}
\resizebox{\linewidth}{!}{%
\begin{tabular}{|c|c|c|c|c|c|}
\hline
\textbf{Variable X} & \textbf{Variable Y} & \textbf{P-Value} & \textbf{\begin{tabular}[c]{@{}c@{}}Effect Size\\ (Cohen's d)\end{tabular}} & \textbf{\begin{tabular}[c]{@{}c@{}}Difference \\ Between Means \\ (Y - X)\end{tabular}} & \textbf{\begin{tabular}[c]{@{}c@{}}95\% confidence\\ interval of difference\end{tabular}} \\ \hline
$\overline{T}$      & $\overline{R}$      & 0.029           & 0.249                                                                      & 0.070                                                                                    & 0.010 to 0.140                                                                              \\ \hline
$T^{sim}$ for MPP   & $R^{sim}$ for MPP   & 0.029           & 0.249                                                                      & -0.070                                                                                   & -0.140 to -0.010                                                                            \\ \hline
$\overline{M}$      & $\overline{R}$      & 0.544            & 0.068                                                                      & -0.020                                                                                   & -0.090 to 0.050                                                                             \\ \hline
$\overline{L}$      & $\overline{R}$      & \textless 0.001        & 0.503                                                                      & 0.18                                                                                    & 0.100 to 0.260                                                                              \\ \hline
$\overline{M}$      & $\overline{T}$      & 0.006          & 0.318                                                                      & -0.090                                                                                   & -0.160 to -0.030                                                                            \\ \hline
$\overline{L}$      & $\overline{T}$      & 0.008          & 0.302                                                                      & 0.110                                                                                    & 0.030 to 0.190                                                                              \\ \hline
\end{tabular}%
}
\vspace{-15pt}
\end{table}


\section{Discussion}
\label{sec:HRI_conclusions}
We have provided crucial insights about human behavior and perception to help design better navigational XAI in~RCE. 
We summarize our findings from the user study. Participants tend to prefer riskier and time urgent paths, while they least prefer safer and time relaxed paths. Participants tend to show risk averse and risk insensitive behavior more often than expected behavior with respect to risk. CPT risk model captures participants' decision making better than CVaR and ER, while ER performs the worst of the three models, in accordance with the theoretical findings from previous work~\cite{AS-SM:21-ral,AS-SM:21-lcss}. On an average the survey responses effectively captured participants' risk and time urgency behavior from the trials. However, they may not be fully indicative of people's risk and time urgency preferences as there was a large standard deviation in the similarity scores. Participants tend to prefer safer paths the least compared to their indicated risk appetite for the survey. There was no significant correlation between participants' risk propensity and time urgency indicated in the respective surveys.

Our study reveals that humans act in a diverse manner in RCE, thus motivating the need to equip XAI with more inclusive risk perception models to foster cogent interaction and integration in such environments. We also revealed that standard questionnaires that measure risk propensity and time-urgency scale of individuals may not be very reliable to consistently predict human decision making in everyday RCE. Thus XAI may need to rely on other techniques such as online learning to understand the human perception of risk for efficient communication.    
    
Our research demonstrates insights about users' preferences for XAI-capable mobile robots that communicate their motion intentions.
Overall, we found that most participants want robots that can explain the rationale behind decision-making, factors they considered, and how those factors were weighed against each other to make navigation decisions. 
Also, many participants envisioned intelligent user interfaces to communicate their motion intentions and for robots to express their motion intention on a Google maps-like interface.
For those few participants that indicated no interest in explainable mobile robots, they envisioned XAI interaction could be useful in case of accidents or situations that require a written report. 
Overall, we found much potential for intelligent user interfaces that enable robots to communicate their risk perception to humans and hope that users find this information useful to enable safe and trustworthy mobile robot systems.

\paragraph*{Limitations and Future Work}
Here we will discuss some of the main limitations of our work and then mention possible future work for improvement. Due to the online nature of the study, participants' path choices may not be fully representative of their actions in real world settings. Also, there might have been an inadvertent sampling bias resulting in a relatively younger age group of a student participant population.  A more in-person study in a real world supermarket or similar setting may be needed to further validate our claims. 

Although, we had a large sample size in terms of number of people, we only collected limited data (only 9 trials) per participant to minimize fatigue. More data points are needed to explicitly compare and characterize decision making between participants and risk models. This limitation can be again alleviated by conducting in-person user studies where data can be collected in a natural and continuous manner (entire paths), which can then be used to perform more rigorous comparisons between various risk models.

We also performed correlation studies (point-biserial correlation) to determine the interaction effects between deviation from expected behavior in MPP and LPP using GRiPS score and time-urgency score. We found no statistical significance in these cases. We believe generic risk and time-urgency questionnaire responses may not reflect the participants' decision making in our particular environment. An in-person study might throw some more light on these interactions. Also, questionnaires focused on navigation in RCE may be needed to better capture human path choices.

\section{Conclusions}
In this work, we found that people tend to exhibit a variety of risk perceptions and behaviors in a crowded social navigation setting. We found that risk models like CPT, that are more expressive and inclusive, can better depict the observed human behavior, which thus support  the previous theoretical findings.
We also found that existing standard questionnaires to determine a users' risk-taking and time-urgency traits were not consistent with the exhibited behavior. In addition, we provided novel insights to consider for future XAI development for social navigation scenarios. 
\bibliographystyle{IEEEtran}
 \bibliography{alias,HRI,SM,SMD-add,references}

\end{document}